\pdfoutput=1

\documentclass[11pt]{article}
\usepackage[x11names]{xcolor}
\usepackage[final]{acl}

\usepackage{times}
\usepackage{latexsym}

\usepackage[T1]{fontenc}

\usepackage[utf8]{inputenc}

\usepackage{microtype}

\usepackage{inconsolata}

\usepackage{booktabs}
\usepackage{graphicx}
\usepackage{multirow}
\usepackage{multicol}
\usepackage{amsfonts}

\usepackage{kotex}
\usepackage{comment}

\usepackage{amsmath}
\usepackage{amsthm}
\usepackage{dsfont}
\usepackage[capitalize]{cleveref}
\crefname{section}{Sec.}{Secs.}
\crefname{table}{Tab.}{Tabs.}
\crefname{equation}{Eq.}{Eqs.}
\crefname{figure}{Fig.}{Figs.}

\usepackage{algorithm}
\usepackage{listings}
\newtheorem*{theorem}{Theorem}

\title{QEFT: Quantization for Efficient Fine-Tuning of LLMs}

\author{Changhun Lee$^{1*}$ \quad Jungyu Jin$^{2*}$ \quad Younghyun Cho$^{2}$ \quad \quad Eunhyeok Park$^{2}$ \\
$^{1}$Department of Convergence IT Engineering \\
$^{2}$Graduate School of Artificial Intelligence \\
Pohang University of Science and Technology (POSTECH) \\
\texttt{\{changhun.lee, jgjin0317, yhcho97, eh.park\}@postech.ac.kr} \\
}

\newcommand{\TIX}{{\mkern-1.5mu\times\mkern-1.5mu}}

\begin{document}
\maketitle
\renewcommand{\thefootnote}{\fnsymbol{footnote}}
\footnotetext[1]{ These authors contributed equally.}
\renewcommand{\thefootnote}{\arabic{footnote}}
\begin{abstract} 
With the rapid growth in the use of fine-tuning for large language models (LLMs), optimizing fine-tuning while keeping inference efficient has become highly important. However, this is a challenging task as it requires improvements in all aspects, including inference speed, fine-tuning speed, memory consumption, and, most importantly, model quality. Previous studies have attempted to achieve this by combining quantization with fine-tuning, but they have failed to enhance all four aspects simultaneously. In this study, we propose a new lightweight technique called Quantization for Efficient Fine-Tuning (QEFT). QEFT accelerates both inference and fine-tuning, is supported by robust theoretical foundations, offers high flexibility, and maintains good hardware compatibility. Our extensive experiments demonstrate that QEFT matches the quality and versatility of full-precision parameter-efficient fine-tuning, while using fewer resources. Our code is available at \url{https://github.com/xvyaward/qeft}.
\end{abstract}

\section{Introduction}

While the outstanding zero-shot performance of large language models (LLMs)~\cite{brown2020language,radford2019language,touvron2023llama,zhang2022opt} significantly contributes to their popularity, their versatility and adaptability are also crucial factors in their widespread adoption. Through transfer learning and fine-tuning, LLMs can be extended to handle unseen or complex tasks, including new data types, which opens up new possibilities across various applications~\cite{qin2023toolllm, hao2024toolkengpt}. As efficient inference and fine-tuning become increasingly important, this paper explores methods to enhance the efficiency of both inference and fine-tuning for LLMs.

Examining past research, several methods have been proposed to enhance the efficiency of inference, including pruning~\cite{frantar2023sparsegpt, sun2023simple}, speculative decoding~\cite{leviathan2023fast, miao2023specinfer}, KV caching~\cite{hooper2024kvquant}, and, particularly, weight quantization~\cite{frantar2022optq, lin2023awq, yao2022zeroquant, shao2023omniquant, lee2024owq}. However, studies focusing on lightweight approaches for fine-tuning remain relatively limited. This is because when considering both fine-tuning and inference, the factors requiring optimization—such as inference speed, training speed, memory consumption, and accuracy—become significantly varied. Balancing all these conditions simultaneously presents a substantial challenge.

\begin{figure}[t]
\centering
\includegraphics[width=\linewidth]{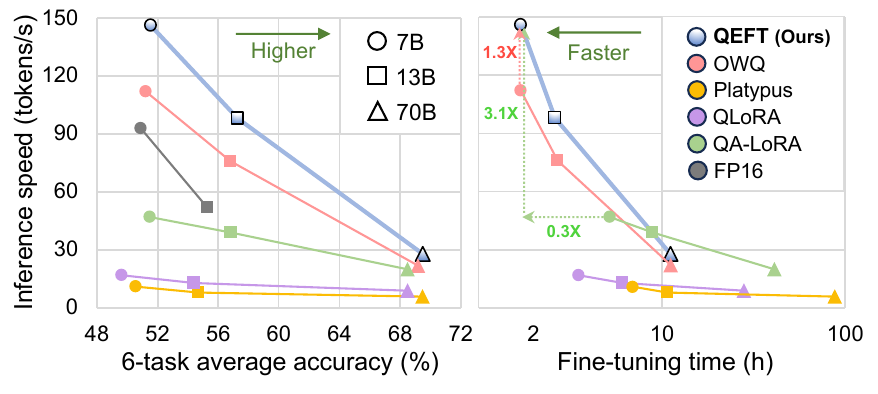}
\caption{Pareto-front comparison of PEFT methods. (Left): Average few-shot accuracy after fine-tuning vs. inference speed. (Right): Fine-tuning time vs. inference speed. For more details, please see \cref{sec:exp_setting}.}
\label{fig:pareto_front}
\end{figure}

In this context, LoRA~\cite{lora2022} is a representative study enabling parameter-efficient fine-tuning (PEFT)~\cite{houlsby2019parameter, zaken2022bitfit, liu2022few, lora2022} by freezing the pre-trained weights while adding a decomposed path that undergoes updates. This approach opens up new opportunities for updating LLMs flexibly with limited resources, leading to numerous emerging applications~\cite{hydra, lee2023platypus, multilora}. In addition, several studies have advanced this concept by attempting to harmonize weight quantization and LoRA to reap the benefits of both methods. For instance, in QLoRA~\cite{dettmers2023qlora} and QA-LoRA~\cite{xu2023qa}, the pre-trained weights remain fixed after quantization, while only the FP16 low-rank path is added and updated exclusively. However, QLoRA exhibits slow inference speeds, and both methods still entail noticeable fine-tuning overhead. Achieving improvements in all aspects of inference and fine-tuning is not easily accomplished by simply applying multiple optimizations in parallel.

In this study, we propose a new quantization technique called Quantization for Efficient Fine-Tuning (QEFT), designed to achieve optimal performance in both inference and training. This method employs the data format of OWQ~\cite{lee2024owq}, storing weak columns vulnerable to quantization in FP16 while storing the majority of weights in 4-bit or less, and updating only the weak columns during fine-tuning. This approach allows us to enjoy the benefits of quantization while implementing PEFT.

However, QEFT offers its own unique innovations. First, OWQ suffers from the irregular mixed precision of columns in the weights of linear layers, resulting in low hardware compatibility. In contrast, QEFT achieves a structured mixed precision representation based on the novel Offline Global Reordering (OGR), improving hardware compatibility and resulting in substantial speed improvements in both training and inference, as shown in \cref{fig:pareto_front}. Additionally, QEFT provides a theoretical framework for selecting weak columns to minimize loss values after fine-tuning. Lastly, despite being implemented differently from LoRA, we validate that QEFT can replace and be applied to applications that previously used LoRA, demonstrating its flexibility.
Through various experiments, we showed that QEFT is the state-of-the-art method in terms of inference speed, training speed, and model quality. While it consumes slightly more memory than OWQ, QEFT outperforms it in every other aspect and surpasses other baselines in all areas.

\section{Related Work}
\subsection{Weight-only Quantization of LLMs}
Weight-only quantization stands out as one of the most successful optimization methods for LLMs, significantly reducing the model's footprint and mitigating memory bottlenecks during generation, thereby notably accelerating inference. OPTQ~\cite{frantar2022optq} pioneered this approach by demonstrating that the OPT-175B model can be quantized to sub-4-bit without notable accuracy degradation. Moreover, this low-precision approach addresses the memory bottleneck, achieving performance benefits on real GPU devices. AWQ~\cite{lin2023awq} and TEQ~\cite{cheng2023teq} make advances that improve the model quality via fine-grained group-wise quantization.

\subsection{Parameter-efficient Fine-tuning (PEFT)}
PEFT is designed to minimize fine-tuning costs for LLMs, unlocking their potential to address new problems affordably. LoRA~\cite{lora2022} is a representative study that freezes pre-trained weights and adds low-rank parameters, which are updated exclusively during fine-tuning. LoRA demonstrates PEFT's potential by showcasing LLMs' remarkable adaptation to unseen tasks with a constrained update scheme.

\subsection{Quantization-aware PEFT}
QLoRA~\cite{dettmers2023qlora} expands upon the LoRA concept by incorporating weight quantization, compressing pretrained weight through quantization. While this makes the fine-tuning process more lightweight, the additional path slows down inference performance, as the additional FP16 path cannot be freely merged to quantized base weights.

QA-LoRA~\cite{xu2023qa} offers an alternative approach where the updated weight is merged on top of the zero-point of the low-precision weight. 
This eliminates the need for a high-precision path after fine-tuning unlike QLoRA, and demonstrates comparable fine-tuning quality; however, it still has a large fine-tuning overhead, and the flexibility for the advanced LoRA applications such as PEFT merging is not explored. 

\subsection{Outlier-aware Weight Quantization}
Recently, OWQ \cite{lee2024owq} introduced intra-layer mixed precision quantization scheme for weight only quantization. In OWQ, the goal is to reduce the layer-wise error, which is broken down into the error for the \(i\)-th output channel as follows:
\begin{equation}
E_i = ||W_{i,:}X - \hat{W}_{i,:}X||_2^2 \approx \Delta W_{i,:} H \Delta W_{i,:}^T.
\label{eq:error_H}
\end{equation}
The Hessian of the weight matrix plays a crucial role in estimating the sensitivity of specific weights. However, the weights in the same output channel share an identical Hessian value, calculated as:
\begin{equation}
H^{(i)} = H = \frac{\partial^2E_{i}}{\partial W_{i,:}^2} = 2XX^T.
\label{eq:hessian}
\end{equation}
From this formulation, it was observed that activation outliers can significantly increase the sensitivity of specific weight columns even when only weight quantization is applied. In OWQ, they calculated the sensitivity of $j$-th column due to weight quantization as follows:
\begin{equation}
sensitivity_j = \lambda_j || \Delta W_{:,j} ||_2^2,
\label{eq:sensitivity}
\end{equation}
where $\lambda$ is a diagonal element of the Hessian. Afterward, they preserved the top-\(k\) most sensitive columns (weak columns) in FP16 and compressed only the remaining robust weights into 4 or 3-bit. 

OWQ enhances the model quality significantly while only adding an extra 0.01 bits on average. However, its mixed-precision format poses challenges in deployment in both GPU and non-GPU environments, limiting thier practical benefits.

\subsection{Weak Column Tuning}
OWQ also introduced the concept that enables PEFT with mixed-precision. This idea, known as Weak Column Tuning (WCT), updates the FP16 weak columns in a task-specific manner while freezing the remaining quantized data. WCT in OWQ sustains the benefits of low precision for both inference and fine-tuning, but it also has several limitations; WCT lacks theoretical support to guarantee its optimality in selecting tuning parameters based on weak columns and has only demonstrated feasibility for specific tasks. Furthermore, the versatility of WCT has not yet been validated, making it less favored than LoRA-based approaches. Most importantly, the irregular mixed-precision weights in OWQ poses challenges for acceleration.

\begin{figure}
\centering
\includegraphics[width=\linewidth]{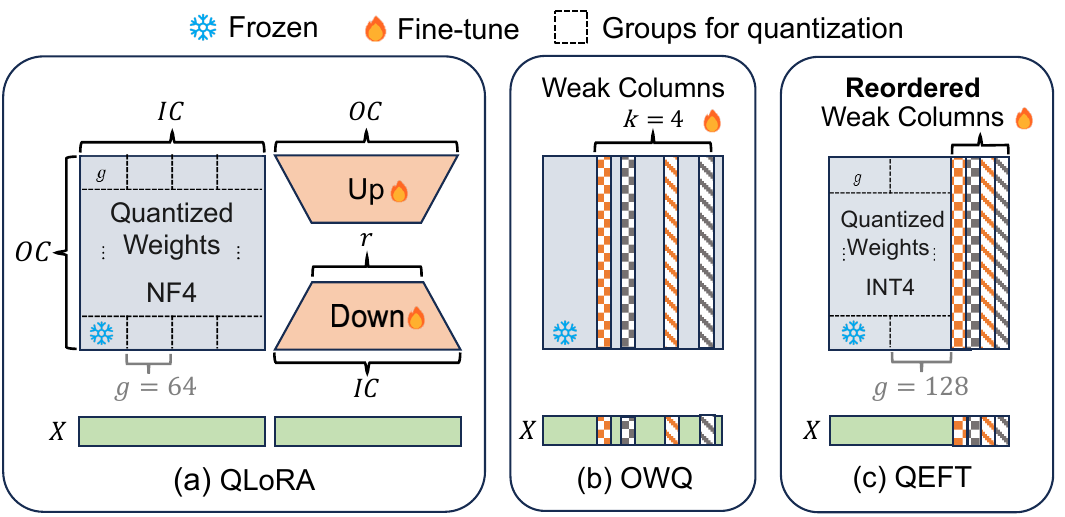}
\caption{The overview of adaptable quantization includes: (a) QLoRA, (b) OWQ, and (c)  the proposed QEFT with OGR. \(k=4\) case as an example.}
\label{fig:wct_main}
\end{figure}

\section{Detailed Overview of QEFT}
\label{sec:QEFT_intro}
In this study, we introduce QEFT, a mixed-precision quantization that achieves higher speed for both inference and fine-tuning and better fine-tuned quality thanks to its hardware-friendly and expressiveness. We begin by detailing QEFT in this section and explain fine-tuning capabilities in \cref{sec:optimal_mask}. Then, we discuss PEFT merging, an advanced example validating its versatility, in \cref{sec:QEFT_merging}.

\subsection{Data Structure and Quantization Process}
\label{qeft_quantization}
QEFT applies mixed-precision quantization to the dense weights of linear layers in LLMs. After quantization, three data components are generated: the dense low-precision matrix, group-wise quantization parameters, and high-precision weak columns, as depicted in \Cref{fig:wct_main}(c). Similar to OWQ, we begin by identifying the top-\( k \) sensitive columns and preserving them in FP16. However, the key difference in implementation is that QEFT employs novel Offline Global Reordering (OGR), as described in \Cref{sec:reordering}, ensuring a structured format unlike to the irregularity of OWQ (\Cref{fig:wct_main}(b)).

Subsequently, the remaining weights are stored in 4-bit or less.
We introduce group-wise quantization~\cite{lin2023awq} to further minimize quantization errors from per-channel quantization based OWQ. Therefore, every adjacent \( g \) weights share the same quantization parameters, such as scaling factor and zero-point. We perform a grid search for each group to find the quantization parameters that minimize the squared error of weights after truncation. Then, we apply OPTQ~\cite{frantar2022optq} using the searched parameters to find the best low-precision mapping. While OPTQ originally relies on channel-wise min-max quantization, ours leverage the benefits of group-wise quantization and truncation~\cite{esser2019learned, librecq, nahshan2021loss, weiqdrop}, resulting in high-quality quantized weights.

\subsection{Offline Global Reordering}\label{sec:reordering}
In OWQ, mixed-precision makes it difficult to accelerate. While the indices of weak columns are predetermined offline, correlated with the location of activation outliers (\Cref{fig:wct_main}(b)), the irregular mixed-precision format introduces multiple branches in the decompression process, causing complex implementation and slowdown. Moreover, these characteristics are difficult to support on emerging hardware, such as in-DRAM accelerators \cite{lee2021hardware} or NPUs \cite{gaudi3_2024}, which only support dense computation in general.

\begin{figure}[t]
\centering
\includegraphics[width=\linewidth]{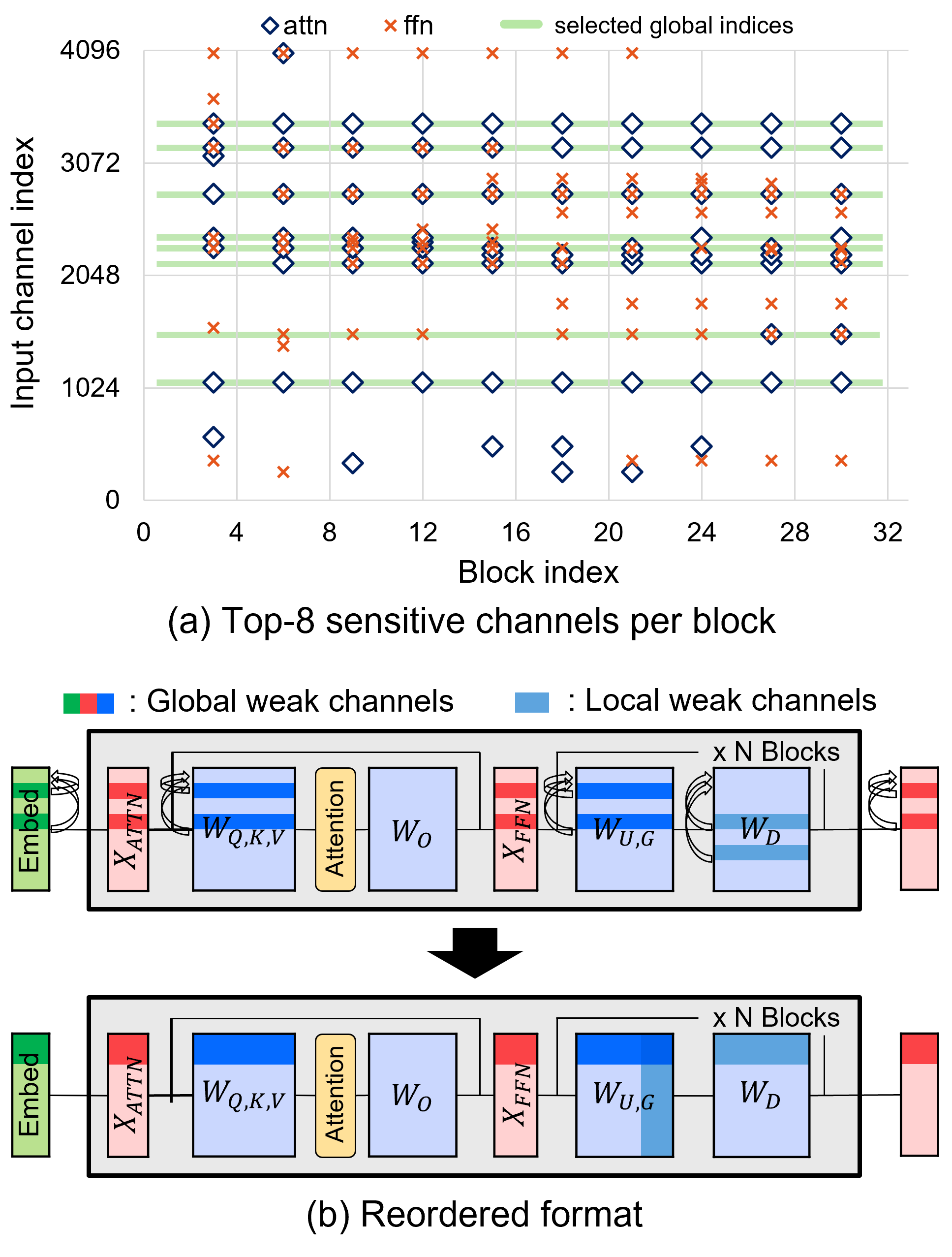}
\caption{(a) Weak column indices at all transformer blocks in the Llama-2 7B model, where "attn" indicates input activation of the attention block and "ffn" indicates input activation of the feed-forward block. (b) An overview of the offline global reordering.}
\label{fig:outlier_indices}
\end{figure}

To address this limitation, we must revise the data representation for predictability and contiguity.
Previous efforts, such as reordering activations in the normalization layer during inference~\cite{yuan2023rptq}, grouped the activation channels in clusters. Although this mitigates the quantization difficulty of the activations, the online reordering incurs additional inference latency, amortizing the benefits of quantization.

To eliminate irregularities without incurring online costs, we introduce a novel concept called Offline Global Reordering (OGR). This idea is motivated by our key observation that outlier channel indices significantly overlap across layers, as shown in \Cref{fig:outlier_indices}(a), which visualizes the layer-wise indices of the top-8 weak columns for each transformer block in the Llama-2 7B model. This overlap occurs because outlier activations propagate through subsequent layers via residual connections, causing weak column indices to align across layers. Based on this observation, we identify and use the common (global) weak columns across the entire network, 
as depicted in \Cref{fig:outlier_indices}(a). Since weight perturbation can vary significantly across layers, we use only $\lambda_j$, instead of $\lambda_j || \Delta W_{:,j} ||_2^2$ in \cref{eq:sensitivity}. The optimality of this metric is discussed in \Cref{sec:optimal_mask}, and the detailed selection algorithm is in \Cref{algo}.

Once the global weak columns are selected, we can rearrange the weights of the embedding and head layers, as well as the layers within the transformer block, offline, as shown in \Cref{fig:outlier_indices}(b). By globally reordering the model, the weak columns of each linear layer form a structured dense matrix, and their corresponding activations are located contiguously. One exception is the attention output projection layer, or $W_O$ in \Cref{fig:outlier_indices}(b). To maintain the multi-head attention mechanism, reordering is not applicable to the $W_O$ weight. In this case, the mixed-precision format is used without reordering.

\Cref{tab:QEFT_ablation} shows the effect of the component proposed in QEFT for Llama-2 7B. For reference, we begin with the OWQ configuration and measure the impact of online reordering and OGR, respectively. The table indicates that online reordering offers limited benefits due to its overhead, whereas OGR significantly accelerates inference.
Although most weak columns overlap, utilizing global weak columns may result in subtle accuracy degradation due to the small number of non-overlapping column indices.
To realize the Pareto-front solution, we seek to address this degradation and consider using group quantization instead of OWQ's channel-wise quantization. A group size of 128 was used by default and it improves the fine-tuned performance by reducing quantization error, with negligible hardware overhead thanks to our optimized kernel. Due to the increased number of group-wise parameters, a slight increase in memory usage occurs from 3923MB to 4107MB. Results show that using OGR with group quantization is the best option and globally reordered model exhibits nearly the same few-shot scores as the optimal selection at each layer while greatly enhancing inference speed.

\begin{table}[t]
    \centering
    \resizebox{\columnwidth}{!}{
    \begin{tabular}{c c|c|c|c}
    \toprule[1.5pt]
        Reorder & Group-wise & 6task & 4task & Inference speed\\
        technique & quantization & Avg. $\uparrow$ & Avg. $\uparrow$ & (tokens/s) $\uparrow$\\
        \midrule[0.75pt] 
         &  & 51.24 & 55.46 & 112 \\
        Online &  & 51.24  & 55.46 & 127 \\
        OGR &  & 51.11 & 55.19 & 148 \\
         & $\checkmark$ & 51.42 & 55.81 & 111 \\
        OGR & $\checkmark$ & 51.55 & 55.70 & 146 \\
        \bottomrule[1.5pt] 
    \end{tabular}
    }
    \caption{Ablation results of QEFT ($k=128$) for the few-shot average scores and inference speed on Llama-2 7B. The bottom row represents QEFT. For more details, please see \cref{sec:exp_setting}.}
    \label{tab:QEFT_ablation}
\end{table}

\subsubsection{GPU Acceleration Kernel for QEFT}\label{sec:custom_kernel}
To realize the full potential of QEFT, we developed a customized matrix-vector multiplication GPU kernel tailored for the reordered format. This kernel first processes the quantized dense matrix by dequantizing the weights into FP16 format and multiplying them with the activations. Subsequently, it performs the multiplication of the high-precision dense weights and activations. Thanks to OGR, we can seamlessly apply two dense computations, which are fused into a single kernel in practice.
The impact of our customized kernel on inference performance is validated in \Cref{inference}.

\begin{figure}[t]
\centering
\includegraphics[width=0.95\linewidth]{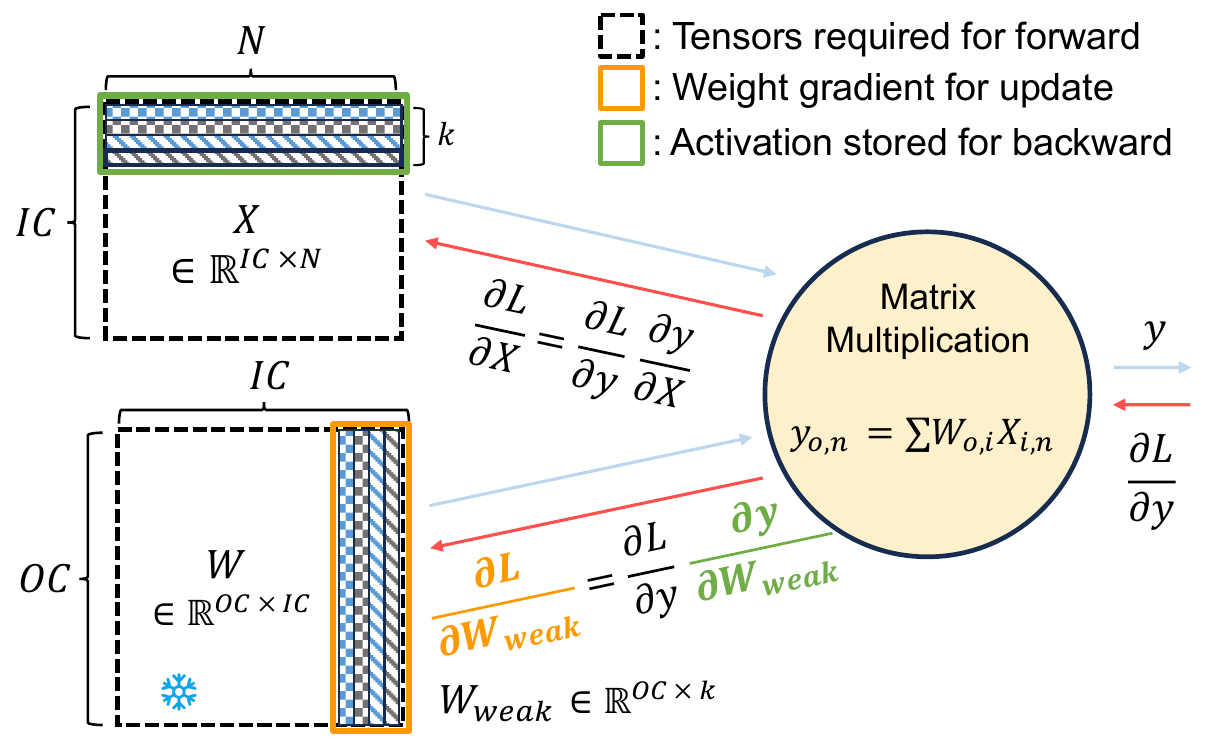}
\caption{Visualization of the reduction in computation and memory usage of QEFT in linear layers.
}
\label{fig:efficient_backward}
\end{figure}

\subsection{Efficient Backward Computation}
\label{sec:efficient_backward}
Utilizing QEFT provides another significant advantage for fine-tuning. As depicted in  \Cref{fig:efficient_backward}, the backward computation of the linear layer involves two GeMM operations: calculating gradients for the input \(X\) and the weights \(W\). Unlike LoRA-based approaches, QEFT reduces the GeMM cost by computing gradients only for the rectangular-shaped trainable weights, thus decreasing the overall FLOPs of weight gradients to \(k/IC\). This reduction offers substantial performance benefits during fine-tuning. Additionally, as shown in  \Cref{fig:efficient_backward}, we only need to store the subset of activations corresponding to the weak columns. The gradient of weights for weak columns can be computed without requiring the entire activation tensor, reducing the memory footprint to \(k/IC\). A crucial aspect of QEFT is that it uses a structural data representation based on OGR, allowing for easy backward implementation in existing frameworks such as PyTorch \cite{paszke2019pytorch}.

\section{Optimal Weak Column Selection}
\label{sec:optimal_mask}
While QEFT is efficient for both fine-tuning and inference, it also offers superior fine-tuning quality.
In this section, we provide theoretical support by proving that selecting weak columns as mask for tunable parameters is an optimal strategy for minimizing the loss value after sparse PEFT, under the following conditions: 1. \textit{A fixed budget is allocated to each linear layer.} 2. \textit{The selection is applied at a per-channel granularity.}

Firstly, we formulate the sparse PEFT as follows:
\begin{equation}
    \min\limits_{\Delta\theta, M} L(\theta^0 + (\Delta\theta)M), 
\end{equation}
where \(\theta^0 \in \mathbb{R}^{OC \times IC}\) represents the pre-trained weights, and \( OC \) and \( IC \) represent the output and input channel dimensions, respectively. \(\Delta\theta \in \mathbb{R}^{OC \times IC}\) represents the updated weights, \(L\) represents the target loss function, and \(M \in \mathbb{R}^{IC\times IC}\) represents the channel-wise parameter mask, where \(M_{i,j} = 0\) if \(i \neq j\) or \(M_{i,i} \in \{0,1\}\) otherwise. To maximize the effect of fine-tuning, we need to select an appropriate \(M\) that can minimize the loss. According to the second-order approximation method of \cite{fu2023effectiveness}, we can find out the optimal mask based on the magnitude of the gradient. 

\begin{theorem}
  \small 
  \begin{equation*}
  \text{if} \;\; \hat{M}_{ii}\!=\!\mathds{1}\Bigg(\sum_{j=1}^{IC}\mathds{1}\bigg(\left|\frac{\nabla L(\theta^0)_{:,i}^2}{h_i}\right| \!>\! \left|\frac{\nabla L(\theta^0)_{:,j}^2}{h_j}\right|\bigg) \!\ge\! IC-k \Bigg), 
  \end{equation*}
   \text{where} $\nabla L(\theta^0)_{:,i}$ \text{is the} $i$\text{-th channel of} $\nabla L(\theta^0)$, \text{then}
   \begin{equation*}
  \inf_{\Delta\theta} L(\theta^0+(\Delta\theta)\hat{M})\le \inf_{\substack{\Delta\theta,\|M\|_0=k;
  }}  L(\theta^0+(\Delta\theta)M).
  \end{equation*}
\end{theorem}

In the constraints of the channel-wise parameter mask, this theorem states that the mask $\hat{M}$ minimizing the infimum of loss can be constructed by selecting the top-$k$ channel indices in order of largest \(|\nabla L(\theta^0)_{:,i}^2|\) values.

In QEFT, tunable weak columns are selected by \cref{eq:sensitivity}, which is based on $\lambda_i$ and the weight perturbation. 
Meanwhile, the gradient of the weight in the linear layer is calculated by the chain rule:
\begin{equation}
\nabla L(\theta)=\frac{\partial L}{\partial \theta}=\frac{\partial L}{\partial y} \frac{\partial y}{\partial \theta}=\frac{\partial L}{\partial y} X^T,
\label{eq:weight_gradient}
\end{equation}
where $X$ represents the activation and $y=\theta X$.
Most importantly, the presence of activation outliers causes both weak column selection (\Cref{eq:sensitivity}) and weight gradient to be dominated by the activation, thus the selection metric of QEFT is also valid for selecting columns with the largest  \(|\nabla L(\theta^0)_{:,i}^2|\).

\begin{figure}[t]
\centering
\includegraphics[width=\linewidth]{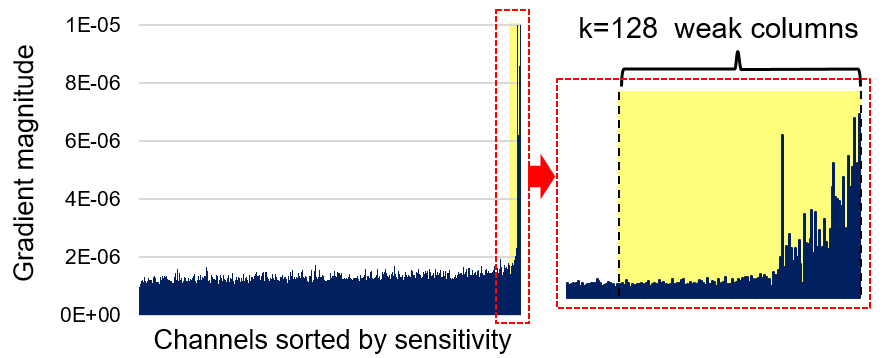}
\caption{Channel-wise sensitivity and the average magnitude of the gradient in Llama-2 7B model. The yellow box indicates the selected weak columns for $k=128$ case.}
\label{fig:gradient_scale}
\end{figure}

\Cref{fig:gradient_scale} illustrates the correlation between \Cref{eq:sensitivity} and \(|\nabla L(\theta^0)_{:,i}^2|\). Sorting the channels using quantization sensitivity reveals that the top-\(k\) channels (weak columns) also represent the columns with the largest gradient magnitude. This implies that although we select weak columns considering the quantization sensitivity, fine-tuning quality is also accounted for.

\section{Advanced Application: PEFT Merging}
\label{sec:QEFT_merging}

\begin{figure}[t]
\centering
\includegraphics[width=0.95\linewidth]{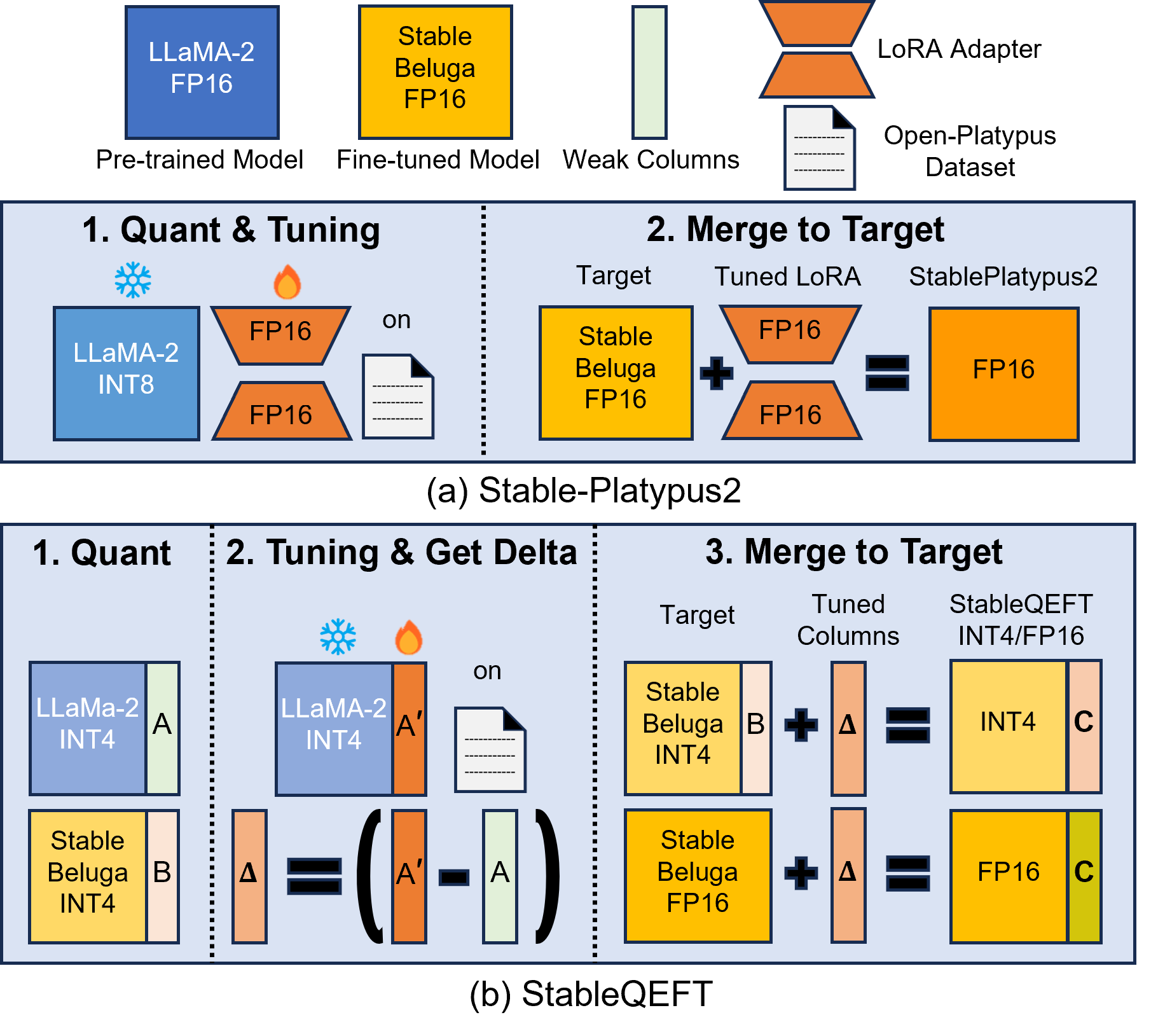}
\caption{An overview of LoRA-based PEFT merging and its QEFT counterparts, QEFT merging. (a) Target model + LoRA case. (b) Target model + QEFT case.}
\label{fig:peft_merging}
\end{figure}

QEFT is designed for efficient inference and fine-tuning, but it also needs to be general enough to be an alternative to the LoRA-based approach. To validate this, we applied QEFT to an advanced application of LoRA known as PEFT merging \cite{lee2023platypus}. \Cref{fig:peft_merging} provides an overview of this process. As shown in the figure, the LoRA adapter is fine-tuned using the Open-Platypus dataset \cite{lee2023platypus} on the Llama-2 model. After fine-tuning, the updated weights are transferred to the StableBeluga \cite{StableBelugaModels} model, which was also initialized with Llama-2 but fine-tuned on a different dataset, resulting in the creation of Stable-Platypus2. Despite potential differences in semantics, Stable-Platypus2 surprisingly exhibits better quality than each model before the merge.

To assess the merging ability of QEFT, we also attempt a similar approach, called QEFT merging. We generate the quantized Llama-2 model and fine-tune the weak columns for the Open-Platypus dataset. The updates of the weak columns ($\Delta = A^{'}-A$) are then merged into the StableBeluga model. If the target model is a full-precision model, we add the update according to the weak column index. For the quantized StableBeluga, we add the update to the corresponding weak columns. 
We show that QEFT merging works surprisingly well in \Cref{sec:peft_merging}, validating its generality.

\renewcommand{\arraystretch}{0.90}
\begin{table*}[hbt!]
    \centering
    \resizebox{\textwidth}{!}{
        \setlength{\tabcolsep}{4pt}
        \begin{tabular}{l || c  c  c || c || c  c||cccccc|| c  c}
            \toprule[1.5pt]
            
            \multirow{2}{*}{\textbf{Model}}  & \multirow{2}{*}{\small{\textbf{Bits}}} & \small{\textbf{Base}} & \small{\textbf{Tunable}} & \textbf{Tuning} & \multicolumn{2}{c||}{\textbf{Inference Speed $\uparrow$}} & \multirow{2}{*}{\small{\textbf{MMLU}}} & \small{\textbf{Hella}} & \multirow{2}{*}{\small{\textbf{ARC-c}}} & \small{\textbf{Truthful}} & \small{\textbf{Wino}} & \multirow{2}{*}{\small{\textbf{GSM8k}}} & \textbf{6task} & \textbf{4task} \\
            
              & & \small{\textbf{Size}} & \small{\textbf{Params.}} & \textbf{Time $\downarrow$} &  \small{\textbf{(tokens/s) }} & \small{\textbf{speedup }} & & \small{\textbf{Swag}} & & {\small{\textbf{QA}}} & \small{\textbf{grande}} & & \textbf{Avg. $\uparrow$} & \textbf{Avg. $\uparrow$} \\
            \midrule[0.75pt] 
            Llama-2 \textbf{7B}  & 16 & 12.9GB & - & - & 93 & 1$\TIX$ &46.54 & 78.63 & 52.99 & 38.96 & 73.64 & 14.63 & 50.90 & 54.28\\
            LoRA  & 16 & 12.9GB & 160M & 1.9h & 93 &1$\TIX$& 48.18 & 78.32 & 55.29 & 41.78 & 74.27 & 1.67 & 49.92 & \underline{55.89} \\
            Platypus  & 8 & 6.7GB & 23M & 7.0h & 11 & 0.12$\TIX$ & 49.93 & 78.74 & 54.44 & 42.70 & 74.79 & 2.84 & 50.57 & \textbf{56.45} \\
            QLoRA   & 4 & 3.7GB & 160M & 3.5h & 17 & 0.18$\TIX$ & 48.44 & 77.75 & 54.44 & 41.71 & 74.12 & 1.48 & 49.65 & 55.58 \\
            QA-LoRA   & 4 & 4.4GB & 89M & 5.2h  & 47 & 0.51$\TIX$ & 48.61 & 78.37 & 53.11 & 41.28 & 73.68 & 14.14 & \underline{51.50} & 55.41 \\
            OWQ \;  ($k$=16)  & 4 & 3.6GB & 22M & 1.7h & 119 & 1.28$\TIX$ & 46.59 & 78.00 & 52.26 & 40.86 & 73.16 & 12.58 & 50.58 & 54.43 \\
            OWQ \;  ($k$=128)  & 4 & 3.9GB & 174M & 1.7h & 112 & 1.20$\TIX$ & 48.24 & 78.01 & 54.14 & 41.47 & 73.32 & 12.25 & 51.24 & 55.46 \\
            \textbf{QEFT} ($k$=16)  & 4 & 3.8GB & 22M & \textbf{1.7h} & \textbf{148} & \textbf{1.59$\TIX$} & 48.70 & 78.21 & 53.54 & 41.96 & 73.29 & 12.17 & \underline{51.31} & 55.60  \\
            \textbf{QEFT} ($k$=128)    & 4 & 4.1GB & 174M & \textbf{1.7h} & \textbf{146} & \textbf{1.57$\TIX$} & 49.02 & 78.21 & 53.80 & 41.77 & 73.56 & 12.93 & \textbf{51.55} & \underline{55.70} \\

            \midrule[0.75pt]
            \midrule[0.75pt]
            Llama-2 \textbf{13B} & 16 & 24.8GB & - & -  & 52 & 1$\TIX$ & 55.42 &	82.19 &	59.64 & 36.90 &	76.09 &	21.38 &	55.27 & 58.54  \\
            LoRA   & 16 & 24.8GB & 250M & 2.7h  & 52 & 1$\TIX$  & 55.59 &	82.24 &	60.92 & 45.64 &	76.44 &	6.79 &	54.60 & \textbf{61.10} \\
            Platypus  & 8 & 12.7GB & 36M & 10.7h & 8 & 0.15$\TIX$ & 56.70 &	82.32 &	60.37 & 42.16 &	75.85 &	10.66 & 54.67 & 60.38 \\
            QLoRA   & 4 & 6.9GB & 250M & 6.1h & 13 & 0.25$\TIX$ & 55.86 &	81.76 &	59.56 & 44.30 &	76.44 &	8.57 & 54.41 & 60.37 \\
            QA-LoRA  & 4 & 8.2GB & 140M & 8.9h & 39 & 0.75$\TIX$ & 56.66 & 81.95 & 61.22 & 41.75 & 76.91 & 22.51 & \underline{56.84} & 60.39 \\
            OWQ \; ($k$=16)  & 4 & 6.7GB & 34M & 2.6h & 80 & 1.54$\TIX$ & 56.03 & 81.81 & 60.32 & 40.81 & 76.00 & 20.32 & 55.88 & 59.74 \\
            OWQ \; ($k$=128)  & 4 & 7.2GB & 273M & 2.7h & 76 & 1.46$\TIX$ & 57.30 &	82.05 &	60.20 & 42.24 &	76.96 &	22.18 & 56.82 & 60.45 \\
            \textbf{QEFT} ($k$=16)  & 4 & 7.1GB  & 34M & \textbf{2.5h} & \textbf{101} & \textbf{1.94$\TIX$} & 56.40 & 81.71 &	61.86 & 42.99 &	76.24 &	23.13 &	\underline{57.05} & \underline{60.74} \\
            \textbf{QEFT} ($k$=128)  & 4 & 7.6GB & 273M & \textbf{2.6h}& \textbf{98} & \textbf{1.88$\TIX$}  &56.80 &	82.01 &	62.33 & 42.46 &	77.51 &	22.56 & \textbf{57.28} & \underline{60.90} \\

            \midrule[0.75pt]
            \midrule[0.75pt]
            Llama-2 \textbf{70B}  & 16 & 131.6GB & - & - & $11^*$ &  1$\TIX$ & 69.83 &	87.33 &	67.32 & 44.92 &	83.74 &	54.06 &	67.87 & 67.35 \\
            Platypus \textbf{$\dagger$}  & 8 & 66.3GB & 141M & 88h & 4 & 0.36$\TIX$ & 70.04 & 87.02 & 70.14 & 51.13 & 83.74 & 54.89 & \underline{69.49} & \underline{69.58} \\
            QLoRA  & 4 & 34.7GB & 828M & 28.1h & 7 & 0.64$\TIX$ & 69.82 & 87.06 & 69.03 & 51.05 & 84.93 &	55.04 & \underline{69.49} & 69.24 \\
            QA-LoRA   & 4 & 41.8GB & 442M & 41.2h & 20 & 1.82 $\TIX$ & 70.20 & 87.32 & 69.50 & 47.69 & 83.94 &	53.18 & 68.51 & 68.54  \\
            OWQ \; ($k$=16)  & 4 & 33.9GB & 107M & 11.0h & 23 & 2.09$\TIX$ & 69.97 & 86.91 & 69.28 & 51.56 & 84.17 & 54.28 & 69.36 & 69.43 \\
            OWQ \; ($k$=128)  & 4 & 35.3GB & 860M & 11.2h & 22 & 2.00$\TIX$ & 70.25 & 86.89 & 70.31 & 50.02 & 84.53 &	53.15 & 69.19 & 69.37 \\
            \textbf{QEFT} ($k$=16) & 4 & 35.8GB  & 107M  &\textbf{10.9h} & \textbf{29} & \textbf{2.64}$\TIX$ & 70.49 & 86.85 & 70.05 & 50.52 &	83.90 &	56.03 &	\textbf{69.64} & \underline{69.48} \\
            \textbf{QEFT} ($k$=128) & 4 & 37.3GB & 860M  & \textbf{11.1h} & \textbf{28} & \textbf{2.55}$\TIX$ & 70.51 &	86.88 &	69.97 & 51.15 & 83.98 & 54.44 & \underline{69.49} & \textbf{69.63} \\

            \bottomrule[1.5pt]        
        \end{tabular}
        }
    \caption{
    Comparison of PEFT methods for various few-shot tasks. The model group is divided into 7B, 13B, and 70B by horizontal double lines. Among the average scores, we \textbf{bold} the best score and \underline{underline} the second and third-best scores. 
    $\dagger$ denotes that the accuracy was measured using an official checkpoint. The tuning time is measured by A100 GPU hours. * denotes using 2 GPUs, as FP16 70B model causes OOM on single A100 80GB GPU.
    }
    \label{tab:main_fewshot_results}
\end{table*}

\section{Experiments}
\subsection{Experiments Setting}\label{sec:exp_setting}
To demonstrate the superiority of QEFT, we conducted extensive analysis. The fine-tuning environment adheres to the setup of a baseline, Platypus \cite{lee2023platypus}. We utilized the Open-Platypus dataset for fine-tuning, which is specially filtered to remove duplicates and redundancy among 11 open-source datasets. Given that the Open-Platypus dataset focuses on STEM and logic question domains, we also adopted their evaluation method, which includes few-shot tasks from the open-llm-leaderboard \cite{open-llm-leaderboard}.

Following the recent version of the leaderboard, we report scores for 6 tasks (MMLU \cite{hendrycks2020measuring}, HellaSwag \cite{zellers2019hellaswag}, ARC-c \cite{clark2018think}, TruthfulQA \cite{lin2022truthfulqa}, Winogrande \cite{sakaguchi2021winogrande}, and GSM8k \cite{cobbe2021training}), along with their average, to assess fine-tuning performance. Additionally, we report the average score of four tasks (MMLU, HellaSwag, ARC-c, and TruthfulQA) used for evaluation in previous studies \cite{lee2023platypus, xu2023qa}. We utilized lm-eval-harness \cite{eval-harness} to measure few-shot accuracy.

We used Adamw optimizer with batch size of 16. We used constant learning rate of 1e-5 and 5e-6 for $k=16$ and 128, respectively. 
We observed the overfitting issue of OPTQ reconstruction in the 70B model, which is also found in the previous work \cite{wu2023zeroquant}. Therefore, we utilized simple round-to-nearest quantization instead of OPTQ for 4-bit Llama-2 70B results. 
Please refer to \Cref{sec:experimenta_details} for the detailed experiment setups.

\subsection{Overall Fine-tuning Results}
In order to compare the superiority of QEFT, we selecte five representative counterparts: LoRA, Platypus, QLoRA, QA-LoRA, and OWQ. Platypus~\cite{lee2023platypus} utilizes 8-bit quantization for the base model and integrates the LoRA module only into the FFNs. QLoRA/QA-LoRA and OWQ are included for quality and performance comparisons using 4-bit quantization. QLoRA/QA-LoRA applies the LoRA module to all linear layers, while OWQ retains \(k\) weak columns as FP16 for all linear layers. It's important to note that \(k/2 \simeq r\) in terms of tunable parameters, as each LoRA module employs two \(d \times r\) adapters.

\renewcommand{\arraystretch}{0.85}
\begin{table*}[t]
    \centering
    \resizebox{\textwidth}{!}{
        \begin{tabular}{l | c |c |cccc|c}
            \toprule[1.5pt]
            
            \textbf{Model} & \textbf{Type} & \textbf{Bits} & \textbf{MMLU} & \textbf{HellaSwag} & \textbf{ARC-c} & \textbf{TruthfulQA} & \textbf{4tasks Avg. $\uparrow$} \\

            \midrule[0.75pt] 
            Llama-2 13B & PT  & 16 & 55.42 & 82.19 & 59.64 & 36.90 & 58.54 \\

            \midrule[0.75pt]
            StableBeluga-13B  & FT & 16 & 57.65 & \underline{82.35} & 61.95 & 49.21 & 62.79 \\
            Stable-Platypus2-13B  & target + LoRA & 16 & 58.15 & 82.31 & \textbf{62.54} & \underline{52.38} & \underline{63.85} \\
            \textbf{Stable-QEFT-13B }  & \multirow{2}{*}{target + QEFT} & 4 & \underline{58.32} & 81.86 & 62.20 & 52.25 & 63.66 \\
            \textbf{Stable-QEFT-13B}  &  & 16 & \textbf{58.97} & \textbf{82.53} & \underline{62.46} & \textbf{53.71} & \textbf{64.42} \\

            \midrule[0.75pt]
            
            OpenOrcaxOpenChat-Preview2  & FT & 16 & 58.52 & 83.09 & 62.63 & 50.57 & 63.70 \\
            OpenOrca-Platypus2-13B  & target + LoRA & 16 & \underline{59.46} & \textbf{83.21} & 62.88 & 52.69 & 64.56 \\
            \textbf{OpenOrca-QEFT-13B} & \multirow{2}{*}{target + QEFT} & 4 & 59.18 & 82.51 & \underline{63.48} & \underline{53.40} & \underline{64.64} \\
            \textbf{OpenOrca-QEFT-13B} &  & 16  & \textbf{59.47} & \underline{83.10} & \textbf{63.74} & \textbf{54.31} & \textbf{65.16} \\
            \bottomrule[1.5pt]        
        \end{tabular}
    }
    \caption{\label{tab:peft_merging}
    PT, FT, and target + X denote pre-trained model, full fine-tuning, and merging of target full fine-tuned model and PEFT module X, respectively.
    Among the PEFT merging results, we \textbf{bold} the best score and \underline{underline} the second-best score.
    }
\end{table*}

We reproduced all results for our control groups (LoRA, Platypus, QLoRA, QA-LoRA, and OWQ), excluding Platypus 70B due to its resource-intensive nature (requiring 8xA100 GPUs) and lengthy training time. Instead, we utilized the officially provided pre-trained weights of Platypus 70B from Huggingface. Since OWQ does not provide tuning code, we implemented it based on our setup. Therefore, the tuning of OWQ was also accelerated by using the QEFT's customized code.
In addition, different from inference, computations are mostly compute-bound matrix multiplication during fine-tuning. In this case, the speed gain of OGR just comes from the simple dequantization process, resulting in negligible benefit ($\sim0.1$h reduction of training time) compared to OWQ.

Experimental results are detailed in \Cref{tab:main_fewshot_results}. QEFT clearly outperforms the other quantization-aware PEFTs in the 13B model. However, Platypus demonstrates the best tuning performance in the 7B model, primarily because accuracy degradation due to quantization is dominant for the smaller model. Yet, when considering the base size (6.7GB vs. 4.1GB), QEFT's performance stands out.
Moreover, even in the 7B case, QEFT with \(k=128\) shows better results than other 4-bit baselines, and in the 13B case, QEFT with just a budget of \(k=16\) exhibits tuning performance that outperforms others. Interestingly, Platypus, LoRA, and QLoRA consistently display low GSM8k scores on 7B and 13B, indicating they might be overly tuned for conventional 4 benchmarks.

Platypus reports that their 70B model fine-tuning required 4 $\times$ A100 GPUs for 22 hours. In contrast, QEFT can be completed in 11 hours with just a single A100 GPU. This represents approximately an 8-fold acceleration in terms of GPU hours, highlighting the memory-time efficiency feature of the proposed method. Despite having only 1/8 of the training cost, the fine-tuned quality of QEFT-based model is comparable to or better than the Platypus-70B, significantly surpassing the baseline Llama-2 70B. Compared to other 4-bit methods, QEFT consistently shows better accuracy. In particular, QEFT has a much lower tuning time than LoRA-based approaches.

\subsection{PEFT Merging Results}
\label{sec:peft_merging}
\Cref{tab:peft_merging} presents the PEFT merging results for two fully fine-tuned models, StableBeluga and OpenOrca \cite{OpenOrca}. Following the objective of merging strategy from the Platypus study, we report the scores of the four tasks (MMLU, HellaSwag, ARC-c, and TruthfulQA) to assess complementary effects. Although QEFT merging only modifies the weight columns corresponding to weak columns of the fully fine-tuned models, all QEFT merging cases increase MMLU, ARC, and TruthfulQA scores together;
thus, merging tuned weak columns enhances both reasoning ability and knowledge capability.  
Furthermore, QEFT merging with a quantized target model (indicated as 4-bits in the table) demonstrates comparable accuracy to FP16 LoRA merging. In this scenario, we benefit from a reduced merged model size and faster inference from the QEFT format. Both merging techniques prove highly beneficial for enhancing model quality. This observation confirms that QEFT aligns well with the idea of PEFT merging.

\subsection{Inference Acceleration}
\label{inference}
In this section, we demonstrate performance benefits of QEFT in auto-regressive generation scenarios. We benchmark inference speed on the Llama families on A100 80GB GPU, using the full-precision model and different tuning methods as a baseline. 
In this experiment, we utilize the optimized multi-head attention kernel and layer normalization kernel of FasterTransformer \cite{nvidia_2023} for all methods except QA-LoRA, as they use AutoGPTQ \cite{autogptq_2024} framework. Please refer to \Cref{genappendix} for details. We report the results as the median of generated tokens per second, generating 256 tokens at batch size 1.
 \begin{figure}
 \centering
 \includegraphics[width=\linewidth]{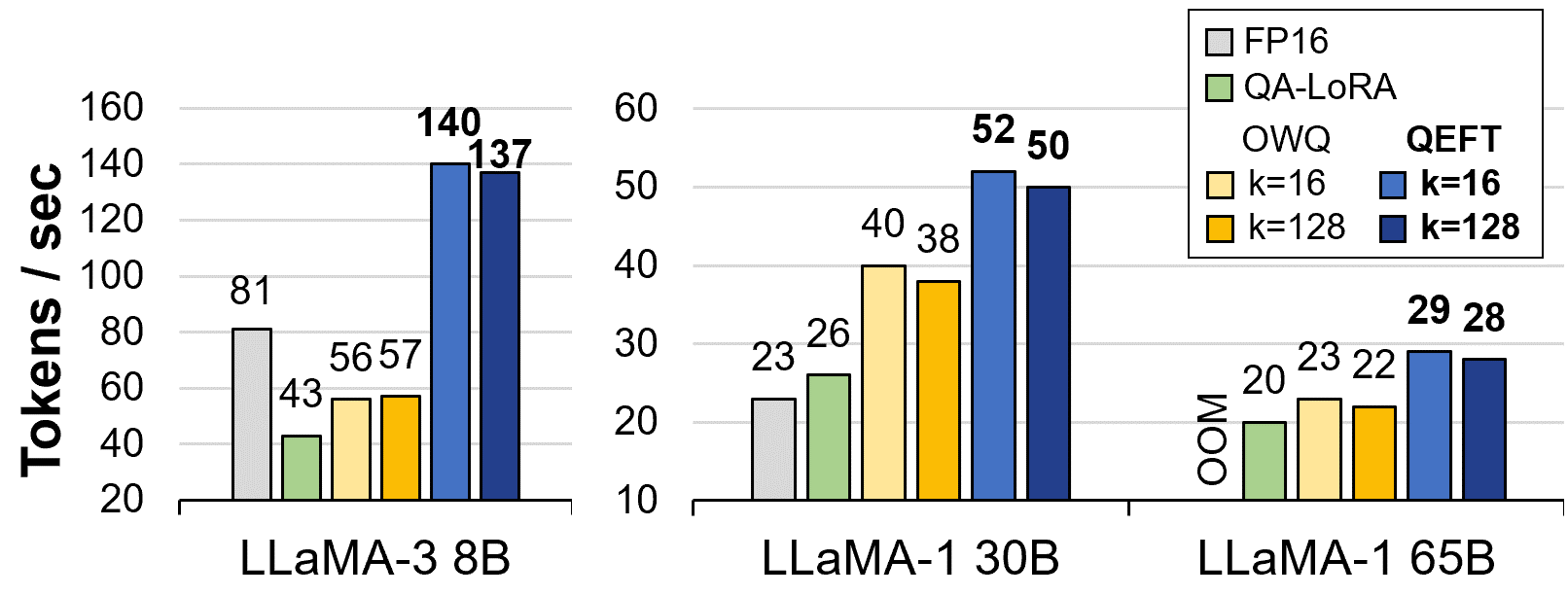}
 \caption{The number of generated tokens per second in auto-regressive generation scenario at the single batch.}
 \label{fig:generation}
 \end{figure}

As shown in the \Cref{tab:main_fewshot_results}, the low-rank path in LoRA-based methods introduces a bottleneck in inference, notably reducing overall speed.
Additionally, as described in \Cref{tab:main_fewshot_results} and \Cref{fig:generation}, even though OWQ also utilizes their optimized kernel, QEFT consistently achieved speedups of about 30\% over OWQ across all model sizes in LLaMA/Llama family. This improvement stems from eliminating the irregular memory access, a major contributor to the overhead of mixed-precision operations, through OGR. In particular, QEFT is about 2.4x faster on the Llama-3 8B model compared to OWQ, because QEFT's kernel exhibits high parallel throughput even for the model with small-sized linear layers and group query attention.

\section{Conclusion}
The storage and computation demands of LLMs present significant barriers to widespread adoption. While quantization and PEFT optimize inference and fine-tuning, respectively, their harmonization is overlooked. In this work, we introduce QEFT, designed to ensure fine-tuning compatibility while prioritizing hardware compatibility. Our experiments confirm that QEFT achieves the fastest fine-tuning and yields the highest few-shot accuracy. Moreover, QEFT shows superior performance in inference, highlighting its excellence across all aspects.

\section*{Limitations}
Although model quantization and parameter-efficient fine-tuning can serve as regularizers during training \cite{askarihemmat2022qreg, fu2023effectiveness}, an explicit regularization mechanism is often beneficial for stabilizing fine-tuning outcomes. In practice, additional regularizers like Dropout \cite{srivastava2014dropout} are employed in LoRA modules to ensure stable training. However, in QEFT, the Dropout-like regularizer is not currently incorporated. While our experiments did not exhibit evident signs of overfitting, this limitation may lead to unaddressed overfitting concerns depending on the specific task. As a future direction, we aim to explore potential candidates for regularization.

In terms of performance, QEFT requires time to convert FP16 to low-precision representation. Typically, the conversion process for the Llama-2 13B model takes about an hour and further details are in the \cref{tab:quantization_cost}. We did not incorporate this conversion cost into \Cref{tab:main_fewshot_results} because it is a one-time expense, and its cost can be amortized by reusing the quantized model for multiple downstream tasks. However, if LLM capacities are further increased, it may be necessary to consider this conversion cost. 

While we can determine the global weak columns and enjoy the benefit of OGR, the `out\_proj' weight cannot be reordered either offline or via equivalent out channel reordering of previous layers. Therefore, we have to specially handle the reordering of `out\_proj' on the fly, utilizing the customized kernel based on OWQ. This hinders the advantage of QEFT's high-throughput generation system.

\section*{Acknowledgement}
This work was supported by Institute of Information \& communications Technology Planning \& Evaluation (IITP) grant funded by the Korea government (MSIT) (RS-2023-00213611, RS-2024-00415602, RS-2024-00396013).

\bibliography{qeft_arxiv}

\appendix

\newpage

\section{Experimental Details}
\label{sec:experimenta_details}

\begin{table*}
    \centering
    \resizebox{\textwidth}{!}{
    \begin{tabular}{c||c|c|c|c|c|c|c|c}
    \toprule
     \multirow{2}{*}{Hyper-parameter}& \multicolumn{2}{c|}{QEFT} & LoRA & Platypus & \multicolumn{2}{c|}{QLoRA} & \multicolumn{2}{c}{QA-LoRA} \\
      \cmidrule{2-9}
         & $k=16$ & $k=128$ & 7B/13B & 7B/13B &  7B/13B & 70B & 7B/13B & 70B \\
    \midrule
        group\_size & \multicolumn{2}{c|}{128} & - & per-tensor & \multicolumn{2}{c|}{64} & \multicolumn{2}{c}{32} \\
        LR & 1e-5 & 5e-6 & 2e-4 & 4e-4 & 2e-4 & 1e-4 & 2e-5 & 1e-5 \\
        Dropout & - & - & 0.1 & 0.05 & 0.1 & 0.05 & 0 & 0 \\
        Scheduler & constant & constant & constant & cosine & constant & constant & constant & constant \\
        warmup steps & 0 & 0 & 0 & 100 & 0 & 0 & 0 & 0 \\
        Low rank (r) & - & - & 64 & 16 & 64 & 64 & 64 & 64\\
        LoRA modules & - & - & all & FFN & all & all & all & all\\
        max\_grad\_norm & 0.3 & 0.3 & 0.3 & 1.0 & 0.3 & 0.3 & 0.3 & 0.3\\
        \cmidrule{2-9}
        per\_device\_train\_batch\_size & \multicolumn{8}{c}{1}\\
        gradient\_accumulation\_steps & \multicolumn{8}{c}{16}\\
    \bottomrule
    \end{tabular}
    }
    \caption{The experimental configurations and hyperparameters for fine-tuning.}
    \label{tab:configuration}
\end{table*}

\subsection{Experimental Details of Fine-tuning}

We used lm-eval-harness \cite{eval-harness} commit version b281b09 for a fair comparison. GSM8k is the only task involving generation among the tasks, so we used a batch size of 1 for evaluating GSM8k to minimize the padding effect. Other few-shot tasks were evaluated using the maximum available batch size. 
We followed the evaluation methods and datasets provided by lm-eval-harness, for example, HellaSwag was evaluated using 10k validation data and GSM8k was evaluated using 1.3k test data.

We utilized gradient checkpointing and gradient accumulation to reduce fine-tuning memory usage. Although gradient checkpointing makes training slower, our tuning costs (GPUh) were measured using these options by default. For QEFT tuning of the Llama-2 70B model, we used the max\_grad\_norm value of 0.0. For the fine-tuning configurations of our control groups, we mostly followed the original training strategy and experimental configurations of each control group. Additionally, the QA-LoRA model is quantized using 128 calibration samples extracted from the C4 dataset. We report the 2 seed average score for 7B/13B in \Cref{tab:main_fewshot_results}. Please refer to the \Cref{tab:configuration} for detailed hyperparameters.

\subsection{Experimental Details of the Generation Benchmark}
\label{genappendix}
QA-LoRA exploits the AutoGPTQ \cite{autogptq_2024} library for their fine-tuning and inference. However, we found that the datatype of AutoGPTQ's zero point is integer while QA-LoRA utilizes floating point zero point. This difference could potentially lead to a decrease in the kernel speed of QA-LoRA when considering full functionality.
QEFT's matrix vector multiplication kernel implementation is based on TensorRT-LLM~\cite{trtllm_2024}.

\section{Algorithm for Global Weak Column Index Selection}
\label{algo}
In the \Cref{sec:reordering}, we analyzed the location of weak columns and proposed efficient reordering by taking advantage of the fact that they overlap a lot. Nevertheless, there are weak columns that do not overlap. To determine optimal global weak column index within limited budget (top-$k <$ union of weak columns), we propose \Cref{alg:code} based on their sensitivity value.  

\begin{algorithm}
\caption{Global index selection}
\label{alg:code}
\lstset{
backgroundcolor=\color{white},
basicstyle=\fontsize{8.5pt}{8.5pt}\ttfamily\selectfont,
columns=fullflexible,
breaklines=true,
captionpos=b,
keywordstyle=\fontsize{9pt}{9pt},
escapeinside={(@}{@)}
}
\begin{lstlisting}[language=python]
# d: hidden state dimension of models.
# k: number of outlier channel
# X: encoded sequences. shape :[b, s, d]

def compute_sensitivity(layer, X):
    H = (2*X @ X.T).mean(dim=0)
    sensitivity = H.diag()
    top_indices = sensitivity.topk(k).indices
    return sensitivity, top_indices
    
s_global = torch.zeros(d)

for block in blocks:
   for layer in block:
      s_local, ids = compute_sensitivity(layer, X)
      s_global[ids] += s_local[ids] / s_local.mean()

global_indices = torch.topk(s_global, k).indices

\end{lstlisting}
\end{algorithm}

\section{Additional Experiments}

\subsection{Few-shot Results on 3-bit Settings}
\label{sec:fewshot_3bit}
We added results for 3-bit settings to \Cref{tab:3bit_fewshot_results}. The accuracy gap between QEFT and other methods increases, except for a single case: QA-LoRA with Llama-2 13B. Two factors mainly affect fine-tuned accuracy: (1) mask selection and corresponding fine-tuning ability, and (2) quality of the frozen quantized weights.

In the case of OWQ and QEFT, the fine-tuning ability of (1) is the same according to the convergence analysis in Section 3. Therefore, the difference in accuracy is caused by (2), where group-wise quantization of QEFT makes a gap in quantization quality. Thus, it is also natural that the accuracy gap increases in 3-bit, where quantization quality is even more important.

On the other hand, QA-LoRA utilizes a group size of 32, so generally QA-LoRA has a better quantization quality regarding (2) while having a large storage overhead of base size in return, as clearly shown in the \Cref{tab:3bit_fewshot_results}. Nevertheless, as QEFT guarantees lower convergence, QEFT shows better accuracy in most 4-bit and 3-bit cases as QEFT's (1) dominates QA-LoRA's (2). However, as quantization quality becomes more important in 3-bit, we assume a single reversed result occurred for 13B.

In this case, the base size of QA-LoRA is 6.68GB, which is about 12\% larger than QEFT's 6.05GB. However, it is difficult to match all other configurations and conditions (e.g., tuning cost, base size, inference speed, etc.).

\begin{table*}[hbt!]
    \centering
    \resizebox{0.75\textwidth}{!}{
        \setlength{\tabcolsep}{4pt}
        \begin{tabular}{l || c | c ||cccccc||c|c ||}
            \toprule[1.5pt]
            
            \multirow{2}{*}{\textbf{Model}} & \small{\textbf{Base}} & \small{\textbf{Tunable}} & \multirow{2}{*}{\small{\textbf{MMLU}}} & \small{\textbf{Hella}} & \multirow{2}{*}{\small{\textbf{ARC-c}}} & \small{\textbf{Truthful}} & \small{\textbf{Wino}} & \multirow{2}{*}{\small{\textbf{GSM8k}}} & \textbf{6task} & \textbf{4task} \\
            
             & \small{\textbf{Size}} & \small{\textbf{Params.}} & & \small{\textbf{Swag}} & & {\small{\textbf{QA}}} & \small{\textbf{grande}} & & \textbf{Avg. $\uparrow$} & \textbf{Avg. $\uparrow$} \\
            \midrule[0.75pt] 
            Llama-2 \textbf{7B} \\
            QA-LoRA & 3.59GB & 89M & 44.97 & 76.68 & 49.91 & 40.75 & 71.98 & 11.90 & 49.36 & 53.08 \\
            OWQ \;  ($k$=128) & 3.15GB & 174M & 45.06 & 76.22 & 52.47 & 39.31 & 72.85 & 10.54 & \underline{49.41} & \underline{53.27} \\
            \textbf{QEFT} ($k$=128) & 3.33GB & 174M & 45.07 & 76.88 & 52.86 & 42.86 & 72.45 & 10.85 & \textbf{50.16} & \textbf{54.42} \\
            
            \midrule[0.75pt]
            Llama-2 \textbf{13B} \\
            QA-LoRA & 6.68GB & 140M & 55.75 & 81.16 & 59.04 & 42.77 & 75.77 & 20.62 & \textbf{55.85} & \textbf{59.68} \\
            OWQ \; ($k$=128) & 5.69GB & 273M & 55.63 & 80.69 &	57.94 & 42.10 &	76.01 &	18.12 & 55.08 & 59.09 \\
            \textbf{QEFT} ($k$=128) & 6.05GB & 273M & 56.18 &	80.67 &	58.41 & 42.54 &	75.89 &	20.74 & \underline{55.74} & \underline{59.45} \\

            \midrule[0.75pt]
            Llama-2 \textbf{70B} \\
            QA-LoRA & 33.64GB & 442M & 69.64 & 86.65 & 68.09 & 46.89 & 83.58 &	52.08 & 67.82 & 67.82 \\
            OWQ \; ($k$=128) & 27.14GB & 860M & 69.05 & 86.08 & 67.92 & 49.81 & 83.66 &	50.72 & \underline{67.87} & \underline{68.22} \\
            \textbf{QEFT} ($k$=128) & 29.10GB & 860M & 70.07 &	85.97 &	68.17 & 52.55 & 83.66 & 51.18 & \textbf{68.60} & \textbf{69.19} \\
            \bottomrule[1.5pt]        
        \end{tabular}
        }
    \caption{\label{tab:3bit_fewshot_results}
    Comparison of PEFT methods for various few-shot tasks on 3-bit base settings. The model group is divided into 7B, 13B, and 70B by horizontal lines. Among the average scores, we \textbf{bold} the best score and \underline{underline} the second scores.
    }
\end{table*}

\subsection{Comparison of Quantization Methods}
Selecting an adequate quantization method is important as it is directly connected to the post-tuned performance. While fine-tuning can boost task-specific accuracy, quantization-aware PEFT must overcome the accuracy degradation from quantization.
We compared two quantization schemes, OWQ and round-to-nearest (RTN) method on Llama-2 7B. In both cases, we preserve $k=8$ weak columns as FP16. OWQ searches clipping value and reduces quantization error by reconstruction, while RTN uses naive minmax quantization. \Cref{tab:quantization_method} shows that few-shot accuracies using OWQ as quantization are consistently better than RTN. Interestingly, perplexity score after quantization (without fine-tuning) is similar in both cases. Analyzing more on the effect of quantization on fine-tuning is our future research direction.

\begin{table}
    \centering
    \resizebox{0.7\columnwidth}{!}{
    \begin{tabular}{ccc|cc}
    \toprule
     & \multicolumn{2}{c|}{WikiText-2 $\downarrow$} & \multicolumn{2}{c}{Avg. of 6 tasks $\uparrow$} \\
         & 7B & 13B & 7B & 13B \\
    \midrule
        RTN & 5.25 & \textbf{4.65} & 50.54 & 56.43 \\
        OWQ & \textbf{5.24} & 4.67 & \textbf{51.16} & \textbf{56.82} \\
    \bottomrule
    \end{tabular}
    }
    \caption{Comparison of quantization methods for the base model.}
    \label{tab:quantization_method}
\end{table}

\subsection{The Number of Weak Columns k}
In this paper, $k$ presents the number of weak columns, which are preserved in full-precision in each linear weight. There is a trade-off depending on the $k$ value between the number of tunable parameters and the overall model cost. To find out the effect of $k$ on tuning performance, we measured the score by changing the value of $k$ (\Cref{tab:k_sweep}). When we used the same learning rate, the score seemed irrelevant to the $k$ values. The results are the average of the three seed values, and although there is some variation, it was confirmed that the performance improves as the k value increases. $k=16$ and 128 were used for the experiments in \Cref{tab:main_fewshot_results} to match the number of learnable parameters to the control groups.

\begin{table}[t]
        \centering
        \small
        \begin{tabular}{ c | c @{\hskip 0.12in} c @{\hskip 0.12in} c @{\hskip 0.12in} c @{\hskip 0.12in} c }
            \toprule
            \textbf{$k$} & \textbf{8}  & \textbf{16} & \textbf{32} & \textbf{64} & \textbf{128} \\
            \midrule
            lr ($\times 10^{-5}$) & 1.4 & 1.0 & 0.7 & 0.5 & 0.4 \\
            \midrule
            \textbf{Avg}& 56.66 & 56.98 & 56.94 & 56.90 & 57.28 \\
            \bottomrule
        \end{tabular}
        \caption{Average score of 6 few-shot tasks according to change in the value of $k$.}
        \label{tab:k_sweep}
\end{table}

\subsection{Quantization Cost for QEFT and OWQ}
We measured quantization time and storage cost for QEFT and OWQ on Llama-2 family (\Cref{tab:quantization_cost}). In the 7B and 13B models, QEFT is more expensive than OWQ because it applies group-wise quantization along with grid search. However, please note that the quantization process time is one-time cost and can be amortized. If we create the quantized weights for each base model once, they can be used for multiple QEFT fine-tuning for several datasets and tasks.
For the Llama-2 70B model, OWQ takes 4.1 hours to perform quantization and reconstruction on the single A100 80GB. On the other hand, group-wise quantization process of QEFT 70B takes about 1.8 hours because QEFT 70B utilizes minmax and round-to-nearest quantization instead of grid search and reconstruction, respectively. Please refer to \Cref{sec:exp_setting} for more details about the quantization of the QEFT 70B case.
Regarding the storage cost, grid search also does not affect the cost as it only alters the value of each quantization parameter (scales and zero points). Group-wise quantization slightly increases storage overhead due to the increased number of group-wise parameters. \Cref{tab:quantization_cost} shows the storage overhead of group-wise quantization for all model sizes (with $k=128$). We can see that group-wise quantization imposes insignificant storage overhead compared to the overall size of the model.

\begin{table}
    \centering
    \resizebox{\columnwidth}{!}{
    \begin{tabular}{ccc|cc}
    \toprule
        & \multicolumn{2}{c}{QEFT} & \multicolumn{2}{c}{OWQ} \\
     \cmidrule{2-5}
        & Time(m) & Storage(MB) & Time(m) & Storage(MB)\\
    \midrule
        7B & 39.5 & 4107 & 25.6 & 3923 \\
        13B & 70.0 & 7561 & 46.7 & 7200 \\
        70B & 107.0 & 37260 & 246.7 & 35296 \\
    \bottomrule
    \end{tabular}
    }
    \caption{Quantization costs for QEFT and OWQ.}
    \label{tab:quantization_cost}
\end{table}

\subsection{Measuring the Tuning Time}
The tuning cost reported in \Cref{tab:main_fewshot_results} measures the time required to train 1 epoch of the Platypus dataset. Please note that these training time results (except Platypus 70B) were measured during the actual process of training QEFT or while reproducing our control group results, under the same environment of a single A100 80GB memory. We referenced the tuning cost of Platypus 70B from its paper due to its resource-intensive nature (requiring 8xA100 GPUs) and lengthy training time. 

Additionally, we compare the forward and backward times for a single model run with an input length of 512. These results are reported in \Cref{tab:forward_backward_time}. Compared to QLoRA, QEFT is significantly faster in both forward and backward passes because there is no low-rank decomposed path. In particular, the backward pass is further accelerated due to the reduced amount of computation, as explained in \Cref{sec:efficient_backward}. When we compare the combined forward and backward time of QEFT and QLoRA, QEFT is more than twice as fast, which is consistent with the overall tuning time of the two methods in \Cref{tab:main_fewshot_results}.

\begin{table}[t]
        \centering
        \small
        \begin{tabular}{ l | c  c | c }
            \toprule
            LLaMa-2 Model & Forward & Backward & Total \\
            \midrule
            QEFT 7B \quad($k$=128) & 54ms & 160ms & 214ms \\
            QLoRA 7B \quad($r$=64) & 138ms & 340ms & 478ms \\
            \midrule
            QEFT 13B \quad($k$=128) & 90ms & 290ms & 380ms \\
            QLoRA 13B \quad($r$=64) & 254ms & 590ms & 844ms \\
            \bottomrule
        \end{tabular}
        \caption{Speed measurement results for forward and backward operations in QEFT and QLoRA. Gradient checkpointing is applied to all methods.}
        \label{tab:forward_backward_time}
\end{table}

\begin{table}
    \centering
    \resizebox{0.7\columnwidth}{!}{
    \begin{tabular}{c|ccc}
    \toprule
      & 7B & 13B & 70B \\
    \midrule
    LoRA & 14.33 & 27.08 & OOM \\
    Platypus & 8.16 & 14.77 & OOM \\
    QLoRA & 5.16 & 9.12 & 41.46 \\
    QA-LoRA & 5.01 & 9.03 & 42.98 \\
    OWQ & 4.10 & 7.41 & 35.75 \\
    QEFT & 4.24 & 7.69 & 37.25 \\
    \bottomrule
    \end{tabular}
    }
    \caption{Training memory footprint (GB) of each PEFT method.}
    \label{tab:training_memory}
\end{table}

\begin{figure}[t]
\centering
\includegraphics[width=0.95\linewidth]{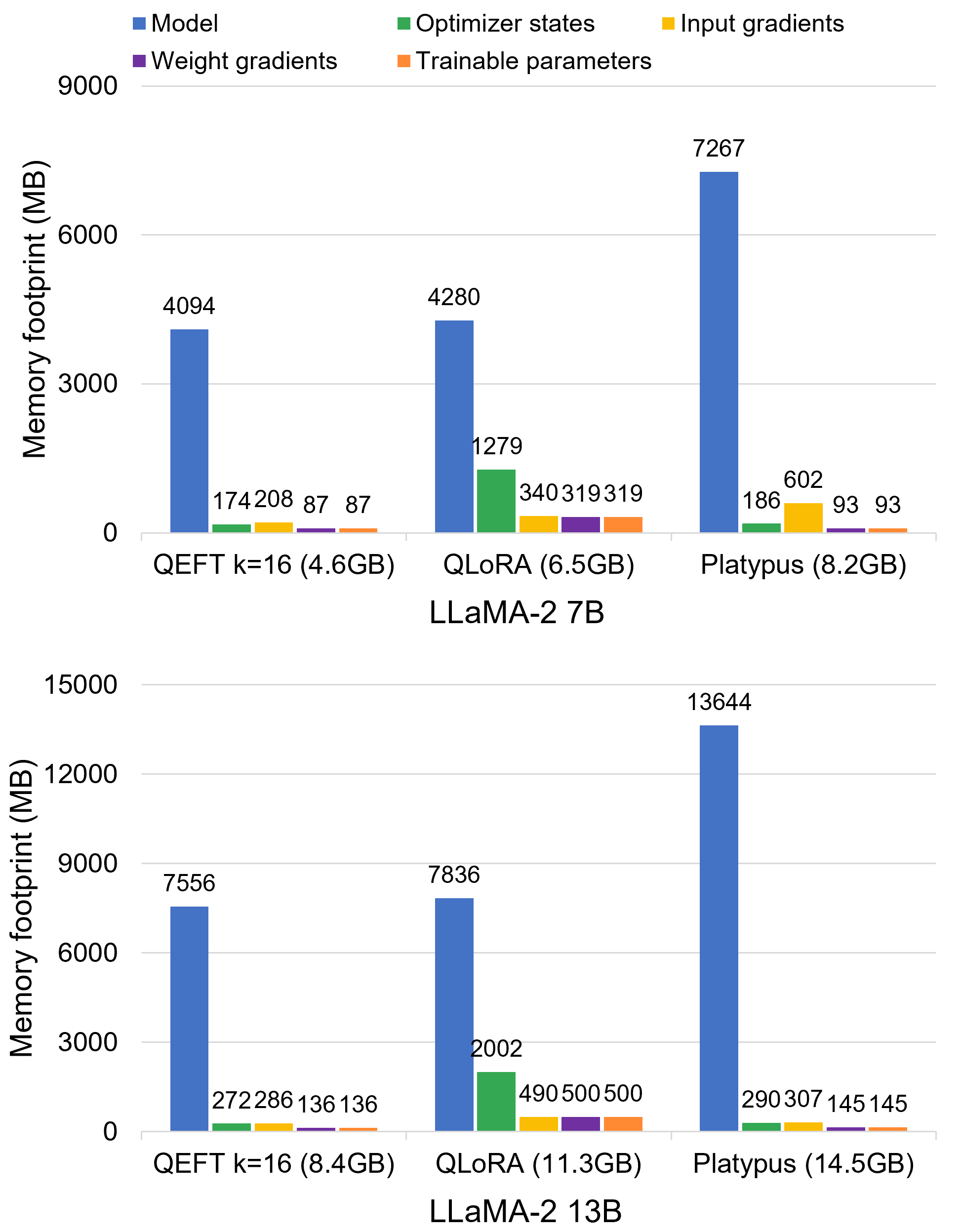}
\caption{The memory usage for training with QEFT and baselines.}
\label{fig:memory_footprint}
\end{figure}

\section{Memory Footprint}
\subsection{Fine-tuning Memory Footprint}

\Cref{tab:training_memory} shows the memory footprint during fine-tuning using each PEFT method. We used the same configuration as \Cref{tab:main_fewshot_results} for all methods, except input sequence length, where we used 512 for this. OWQ and QEFT use $k=16$.

First, the training memory footprint is dominated by the base model size. LoRA has the largest memory footprint because it uses a 16-bit base model, followed by Platypus with an 8-bit base model. Please note that as OWQ does not provide tuning code, we implemented it based on our setup. Therefore, OWQ also benefits from QEFT's customized backward implementation, while both memory footprints are much lower than other methods. The slight difference in memory footprint between OWQ and QEFT is due to the use of groups.

We further detailed a breakdown of memory usage during training in \Cref{fig:memory_footprint}.  
The memory used by the input-enabled gradient in all methods is small because gradient checkpointing is applied, and the batch size is set to 1. At the $k=16$ setting, QEFT incurs the smallest memory usage among QEFT, QLoRA, and Platypus.

\subsection{GPU Memory Overhead of the Quantization process}

\begin{table}
    \centering
    \resizebox{0.9\columnwidth}{!}{
    \begin{tabular}{c|cc}
    \toprule
     Llama-2 & unoptimized (GB) & optimized (GB) \\
    \midrule
    7B & 14.40 & \textbf{6.40} \\
    13B & 18.38 & \textbf{8.38} \\
    70B & 32.51 & \textbf{16.51} \\
    \bottomrule
    \end{tabular}
    }
    \caption{Peak GPU Memory Usage during quantization process of QEFT (and other OPTQ-based methods).}
    \label{tab:quantization_memory}
\end{table}

Additionally, we report the peak GPU memory usage during the QEFT quantization process in \Cref{tab:quantization_memory}. 

First, the memory usage of the quantization process in QEFT, OWQ, and QA-LoRA follows the block-wise reconstruction method of OPTQ, where only each decoder block is kept in GPU memory every time. Therefore, the memory consumption of these methods is almost identical. When we use OPTQ's implementation, the forward process of the calibration set between each block consumes the most GPU memory usage, as it keeps both 128 input and 128 output tensors on GPU memory. This requirement is identical for QEFT, OPTQ, OWQ, and QA-LoRA, demanding a reasonable amount of GPU memory, as seen in the "unoptimized" column of the Table.

However, in our experiment, we further optimized the original OPTQ reconstruction process to lower peak GPU memory usage. By allowing only one calibration sample to be allocated to GPU memory at a time on demand, without affecting the quantization process, we reduced memory consumption by half ("optimized" column of the Table) over the original OPTQ implementation. Please note that all of OPTQ-related algorithms benefit from this optimization as well. In environments where GPU memory for parameter-efficient fine-tuning is available, the base model can be quantized without difficulty.

\end{document}